\documentclass{article}
\usepackage{spconf,amsmath,graphicx,hyperref}
\usepackage{amssymb,bm}
\usepackage{booktabs}
\usepackage{multirow}
\usepackage{array}
\usepackage{xcolor}
\usepackage{placeins}
\usepackage{makecell}

\usepackage{float}
\usepackage{cite}

\makeatletter
\let\oldthebibliography\thebibliography
\def\thebibliography#1{%
  \oldthebibliography{#1}%
  \small 
  \setlength{\parskip}{0pt}%
  \setlength{\itemsep}{0.1pt plus .2pt}%
  \setlength{\parsep}{0pt}%
  \setlength{\labelsep}{0.2em}%
}
\makeatother

\title{Mind the Shift: Using Delta SSL Embeddings to Enhance Child ASR}
\name{Zilai Wang, 
Natarajan Balaji Shankar, 
Kaiyuan Zhang,
Zihan Wang,
Abeer Alwan}
\address{University of California, Los Angeles\\
Department of Electrical and Computer Engineering\\
Los Angeles, USA
}
\begin{document}
\ninept
\maketitle

\begin{abstract}
Self-supervised learning (SSL) models have achieved impressive results across many speech tasks, yet child automatic speech recognition (ASR) remains challenging due to limited data and pretraining domain mismatch. Fine-tuning SSL models on child speech induces shifts in the representation space. We hypothesize that delta SSL embeddings, the differences between embeddings from a fine-tuned model and those from its pre-trained counterpart, encode task-specific information that complements fine-tuned features from another SSL model. We evaluate multiple fusion strategies on the MyST children’s corpus with different models. Results show that delta embedding fusion with WavLM yields up to a 10\% relative WER reduction for HuBERT 
and a 4.4\% reduction for W2V2, compared to fine-tuned embedding fusion. Notably, fusing WavLM with delta W2V2 embeddings achieves a WER of 9.64, setting a new state-of-the-art among SSL models on the MyST corpus. These findings demonstrate the effectiveness of delta embeddings and highlight feature fusion as a promising direction for advancing child ASR.

\end{abstract}

\section{Introduction}
\label{sec:introduction}

Self-supervised learning (SSL) has become a cornerstone of modern speech processing. 
Models pre-trained on large amounts of unlabeled audio routinely achieve impressive performance on automatic speech recognition (ASR) after fine-tuning with a limited set of transcriptions \cite{mohamed2022self}. 
Nevertheless, SSL models continue to face difficulties with child ASR, where word error rates (WERs) remain substantially higher than those for adult speech \cite{fan2024benchmarking}. 
This gap is largely attributed to acoustic and linguistic differences in children’s speech such as higher fundamental and formant frequencies, greater speaker variability, and rapid developmental changes \cite{gerosa2006analyzing,lee1999acoustics,yeung2018difficulties,dutta2022challenges}. 

Recent supervised models like Whisper \cite{radford2023robust} have shown good performance on child ASR \cite{fan2024benchmarking}, in part due to their large-scale training data.
However, these models often function as black boxes, with limited interpretability and less transparent training data, which hinder analysis and raise concerns around reproducibility, data leakage, and bias. In contrast, SSL models remain highly valuable: their embeddings are more interpretable, more transferable to downstream tasks, and are typically trained on open datasets. These advantages motivate continued research on SSL models as a reliable foundation for advancing child speech technologies \cite{ying2025benchmarking,attia2025cpt,carvalho2025ac,sinha2025beyond,horii2025children,medin2024self,li2023towards,sinha2025can}, and underscore the importance of developing methods to further improve their performance.

A diverse set of SSL encoders has emerged, including Wav2vec 2.0 \cite{baevski2020wav2vec}, HuBERT \cite{hsu2021hubert}, WavLM \cite{chen2022wavlm}, and Data2vec \cite{baevski2022data2vec}. These models differ in pre-training objectives and architectures, and thus may yield complementary representations. This has motivated work on feature fusion, where embeddings from multiple SSL models are combined in the representation space. While prior studies have demonstrated the benefits of fusing pre-trained embeddings \cite{tang2022exploring,chen2022fearless,srivastava2023effuse,chiu2024learnable,arunkumar2022investigation}, the potential of fusion between fine-tuned models remains underexplored, particularly for child ASR.

In parallel, recent research on model merging for ASR shows that parameter differences between fine-tuned and pre-trained models (“task vectors”) encode task-specific knowledge and can transfer across domains \cite{ramesh2024task,shankar2025selective,ducorroy2025robust}. However, most model merging approaches assume all constituent models to be fine-tuned from the same pre-trained checkpoint, limiting their ability to exploit the complementary strengths of heterogeneous SSL encoders. Nevertheless, this insight suggests that differences themselves carry rich information about adaptation. 

In this paper, we extend the task vector concept from parameters to representations. We define delta SSL embeddings as the difference between embeddings produced by a fine-tuned model and those of its pre-trained counterpart. This formulation differs from prior work in NLP~\cite{zhang2019delta}, where delta embeddings are introduced as trainable adjustments to frozen word embeddings to mitigate overfitting during fine-tuning. Despite these technical differences, both approaches share the same intuition: deltas capture task-specific information as deviations introduced by fine-tuning relative to the pre-trained reference.
By incorporating delta embeddings into fusion, we aim to highlight the task-specific information and enhance the complementarity between SSL models. Our work thus extends the notion of delta embeddings from word-level regularization in NLP to representation-level fusion of speech signals, broadening their applicability to low-resource ASR.

\textbf{Our contributions can be summarized as follows:}
\begin{itemize}
  \item[\textbf{1.}] We present, to our knowledge, the first study to fuse fine-tuned SSL model representations using delta embeddings for ASR. As a case study, we focus on child ASR and show that incorporating delta embeddings yields gains across heterogeneous encoders. 
  
  \item[\textbf{2.}] On the MyST corpus, we demonstrate that concatenation consistently outperforms weighted sum and cross attention fusion strategies, particularly in few-shot scenarios. This approach achieves a new state-of-the-art WER of 9.64 among SSL models on the MyST corpus.
  
  \item[\textbf{3.}]We show that delta embedding fusion yields strong gains in the extremely low-resource 1\,h setting, achieving a 10\% relative WER reduction for HuBERT and a 4.4\% reduction for W2V2, compared to fine-tuned embedding fusion.

  \item[\textbf{4.}] We employ Canonical Correlation Analysis (CCA) to investigate how delta embeddings enhance complementarity during fusion, offering insights into their effectiveness. \footnote{https://github.com/Zilai-WANG/Delta-Embedding-Fusion}

\end{itemize}

\label{sec:methods}
\section{Methods}
\subsection{Feature Extraction}

We propose fusing fine-tuned embeddings from a reference SSL model with 
delta (\(\Delta)\) embeddings derived from other SSL models. 
We consider three representative SSL encoders that differ in their 
pre-training objectives and modeling strategies:

\begin{itemize}
    \item \textbf{Wav2Vec~2.0} \cite{baevski2020wav2vec}:  
    Learns contextualized speech representations by masking audio frames and predicting their quantized latent codes using a contrastive loss, which distinguishes true codes from distractors.

    \item \textbf{HuBERT} \cite{hsu2021hubert}:  
    Replaces Wav2Vec~2.0’s contrastive prediction with an offline clustering step to generate pseudo-labels.  
    The model is trained to predict labels for masked frames, emphasizing phonetic structure.

    \item \textbf{WavLM} \cite{chen2022wavlm}:  
    Extends HuBERT with gated relative position bias and multitask objectives, combining masked prediction for noise robustness and multi-speaker conditions.
\end{itemize}

Let $f_i$ denote the $i$-th SSL model, with $\mathbf{E}^{(i)}_{\text{ft}}$ and $\mathbf{E}^{(i)}_{\text{pt}}$ denoting embeddings from the fine-tuned and pre-trained models, respectively.  
We define the delta embedding of model $f_i$ as
\begin{equation}
\Delta f_i : \quad 
\mathbf{E}^{(i)}_{\Delta} = \mathbf{E}^{(i)}_{\text{ft}} - \mathbf{E}^{(i)}_{\text{pt}},
\label{eq:delta_embedding}
\end{equation}
Intuitively, the delta embeddings emphasize the changes required for the model to adapt to the downstream task. In this paper, we focus on fine-tuning the pre-trained SSL models for ASR, which necessitates the addition of a CTC layer on top of the pre-trained transformer encoder, and the task-induced shift of model parameters.


\subsection{Feature Fusion Techniques}

We evaluate three strategies for fusing fine-tuned embeddings from one SSL model with delta embeddings from another.  
Let $\mathbf{E}^{(i)}_{\text{ft}} \in \mathbb{R}^{T \times d}$ denote fine-tuned embeddings from model $i$, and $\mathbf{E}^{(j)}_{\Delta} \in \mathbb{R}^{T \times d}$ denote delta embeddings from model $j$, where $T$ is the number of frames and $d$ is the shared feature dimension.  

\begin{itemize}
    \item \textbf{Concatenation (Concat):}  
    Features are concatenated along the feature dimension at each time step:
    \begin{equation}
        \mathbf{Z} = \big[\,\mathbf{E}^{(i)}_{\text{ft}} \,;\, \mathbf{E}^{(j)}_{\Delta}\,\big],
    \end{equation}
    where $[\,\cdot\,;\,\cdot\,]$ denotes concatenation.

    \item \textbf{Weighted Combination (Weighted):}  
    A convex combination parameterized by a learnable scalar $\lambda \in [0,1]$:
    \begin{equation}
        \mathbf{Z} = \lambda \,\mathbf{E}^{(i)}_{\text{ft}} + (1-\lambda)\,\mathbf{E}^{(j)}_{\Delta}.
    \end{equation}

    \item \textbf{Cross-Attention (X-Attn):}  
    To capture complementary information, cross-attention is applied with $\mathbf{E}^{(i)}_{\text{ft}}$ as the query and $\mathbf{E}^{(j)}_{\Delta}$ as both key and value:
    \begin{equation}
        \mathbf{A} = \text{Attention}\!\left(\mathbf{E}^{(i)}_{\text{ft}},\, \mathbf{E}^{(j)}_{\Delta},\, \mathbf{E}^{(j)}_{\Delta}\right),
    \end{equation}
    where $\text{Attention}$ denotes a standard multi-head attention mechanism.  
    The fused representation is obtained as:
    \begin{equation}
        \mathbf{Z} = \text{LayerNorm}\!\left(\mathbf{E}^{(i)}_{\text{ft}} + \mathbf{A}\right).
    \end{equation}
\end{itemize}

\subsection{Canonical Correlation Analysis}

Following prior work \cite{pasad2024self,pasad2023comparative,pasad2021layer,huo2025iterative}, 
we use CCA \cite{hotelling1992relations} 
to quantify similarity between SSL representations.  
CCA measures relationships between two random vectors by identifying the 
linear projections that maximize their correlation.  

Formally, given $n$ paired samples 
$\{(x_1,y_1),\dots,(x_n,y_n)\}$ from 
$X \in \mathbb{R}^{d_1}$ and $Y \in \mathbb{R}^{d_2}$, 
CCA solves
\begin{equation}
v_1, w_1 = \arg\max_{v,w} \; \text{corr}(v^\top X, \; w^\top Y).
\end{equation}
Subsequent projection pairs $(v_i, w_i)$ are chosen to maximize correlation 
while remaining uncorrelated with earlier pairs, up to 
$k = \min(d_1, d_2)$.  
The overall CCA similarity is the average of the canonical correlations 
$\rho_i = \text{corr}(v_i^\top X, w_i^\top Y)$.  

Projection-Weighted CCA (PWCCA) \cite{morcos2018insights} extends it by computing a weighted sum of $\rho_i$, yielding a single scalar similarity score. This measure provides a more robust characterization of representational alignment between multivariate data series. In this work, we employ PWCCA to quantify CCA similarity.



\subsection{Mixture of Experts}
\label{sec:moe}

To further interpret complementarity, we adopt a mixture-of-experts (MoE) gating framework \cite{jacobs1991moe}, following prior work in speech representation fusion \cite{berrebbi2022combining}.  
In this setup, the model produces a weighted combination of fine-tuned and delta embeddings at each frame, where the weights represent the relative contribution of each expert. This approach provides interpretability by explicitly revealing the contribution of fine-tuned and delta embeddings at each frame.

We treat $f_{\text{ft}}$ and $f_{\Delta}$ as the two experts, with the fine-tuned embeddings $f_{\text{ft}}(x)$ serving as input to the gating network.  
The gating weights are computed frame-wise as
\begin{equation}
w(x)_t = \text{Softmax}\!\left(f_{\text{ft}}(x)_t W_{\text{MoE}}\right), 
\quad t = 1,\dots,T,
\end{equation}
where $w(x) \in \mathbb{R}^{T \times 2}$ is the gating matrix and $W_{\text{MoE}} \in \mathbb{R}^{d \times 2}$ is a learnable parameter, with $d$ denoting the embedding dimension.  
We denote the two columns of $w(x)$ as $w_{\text{ft}}(x)$ and $w_{\Delta}(x)$, which provide the gating weights for the fine-tuned and delta embeddings, respectively, each in $\mathbb{R}^{T}$.

The final fused representation at each frame is given by
\begin{equation}
f_{\text{FUSE}}(x)_t 
= \sum_{i \in \{\text{ft}, \Delta\}} w_i(x)_t \, f_i(x)_t, 
\quad t = 1,\dots,T.
\end{equation}


\section{Experimental Setup}
\label{sec:experiments}

\subsection{Datasets}

The My Science Tutor (MyST) corpus \cite{ward2011my} contains roughly 240\,h of transcribed conversational speech from children in grades 3--5, collected during virtual tutoring sessions across science topics.  
Following the protocol of \cite{attia2023kid}, we filter utterances using Whisper-large-v2 transcripts: removing samples with WER~$>50\%$, fewer than three words, or duration~$>30\,\text{s}$. This yields  133\,h train, 21\,h dev, and 25\,h test data. 
To investigate extremely low-resource scenarios, we also construct 1\,h, 5\,h, and 10\,h training subsets.

\subsection{Models}
We evaluate three SSL models available on Hugging Face \cite{wolf2020transformers}: 
Wav2Vec2-Large-lv60 \cite{baevski2020wav2vec}, HuBERT-Large-ll60k \cite{hsu2021hubert}, 
and WavLM-Large \cite{chen2022wavlm}. All models take 16\,kHz waveforms as input and use a convolutional feature encoder that outputs latent representations with an effective stride of 20\,ms and a receptive field of about 25\,ms. These are then processed by a 24-layer Transformer stack with 1024 hidden dimensions. 
Each model is fine-tuned with a character-level CTC loss on MyST 
(full, 10~h, 5~h, and 1~h) following the protocol of \cite{fan2024benchmarking}. 
We refer to them as W2V2, HuBERT, and WavLM, respectively.


\subsection{Feature Fusion}

The upper CTC layer is removed from the fine-tuned model in our framework. Since WavLM achieves the strongest single-model results on MyST \cite{fan2024benchmarking}, we use it as the reference. We fuse its fine-tuned last-layer embeddings with delta embeddings from HuBERT or W2V2.  
We select the last layer as prior analyses show that fine-tuning induces the largest representation shifts in upper transformer blocks~\cite{pasad2021layer}, and the CTC objective primarily optimizes the final layer for ASR.  
During fusion, both WavLM and delta features are frozen, and a new linear CTC head is trained.

We focus on feature-level fusion to evaluate the efficacy of delta embeddings. We benchmark our method against alternative fusion strategies (Table~\ref{tab:fusion_results_myst}) and standard fine-tuned SSL baselines (Table~\ref{tab:ft_baselines}). Furthermore, we provide a direct comparison between full and delta feature fusion across varying training data regimes (Table~\ref{tab:ft_fusion_concat}).




\section{Results}
\label{sec:results}

\subsection{Do delta embeddings improve fusion performance?}

\subsubsection{Comparison of Fusion Methods}
We first compare weighted sum, concatenation, and cross attention across the 1\,h, 5\,h, 10\,h, and full MyST training subsets.  
Two scenarios are considered: (i) fusion of models with similar pretraining procedures (WavLM and HuBERT), and (ii) fusion of models with more diverse pretraining objectives (WavLM and W2V2).  

As shown in Table~\ref{tab:fusion_results_myst}, concatenation consistently outperforms the other methods. We hypothesize that weighted sum is too simple to fully exploit complementary information, while cross-attention tends to overfit in low-resource cases. Concatenation strikes a balance between simplicity and effectiveness, yielding the most consistent improvements.  

In addition, concatenating WavLM with $\Delta$W2V2 generally performs on par with $\Delta$HuBERT in the full, 10\,h, and 5\,h settings, and achieves a clear advantage in the 1\,h case. This suggests that WavLM and HuBERT, which share similar pre-training objectives, produce more redundant representations, whereas W2V2 introduces stronger complementary cues, in extremely low-resource scenarios.

\begin{table}[!t]
\centering
\vspace{-6pt}
\caption{WERs for fusing fine-tuned WavLM with $\Delta$HuBERT or $\Delta$W2V2. Bold indicates the best result for each training size.}
\setlength{\tabcolsep}{4pt}
\resizebox{\columnwidth}{!}{
\begin{tabular}{lllcc}
\toprule
\makecell{\textbf{FT Embed}} & 
\makecell{\textbf{Delta Embed}} & 
\makecell{\textbf{Fusion Method}} & 
\makecell{\textbf{WER}} & 
\makecell{\textbf{Training Size}} \\
\midrule

\multirow{6}{*}{WavLM}
  & \multirow{3}{*}{$\Delta$HuBERT}
      & Weighted & 9.79  & \multirow{6}{*}{Full} \\
  &   & X-Attn   & 10.28 & \\
  &   & Concat   & 9.71  & \\
\cmidrule(lr){2-4}
  & \multirow{3}{*}{$\Delta$W2V2}
      & Weighted & 9.75 & \\
  &   & X-Attn   & 9.80  & \\
  &   & \textbf{Concat} & \textbf{9.64} & \\
\midrule

\multirow{6}{*}{WavLM}
  & \multirow{3}{*}{$\Delta$HuBERT}
      & Weighted & 11.70 & \multirow{6}{*}{10h} \\
  &   & X-Attn   & 12.64 & \\
  &   & \textbf{Concat} & \textbf{11.57} & \\
\cmidrule(lr){2-4}
  & \multirow{3}{*}{$\Delta$W2V2}
      & Weighted & 11.73 & \\
  &   & X-Attn   & 12.96 & \\
  &   & Concat   & 11.61 & \\
\midrule

\multirow{6}{*}{WavLM}
  & \multirow{3}{*}{$\Delta$HuBERT}
      & Weighted & 13.65 & \multirow{6}{*}{5h} \\
  &   & X-Attn   & 14.32 & \\
  &   & Concat   & 12.96 & \\
\cmidrule(lr){2-4}
  & \multirow{3}{*}{$\Delta$W2V2}
      & Weighted & 14.77 & \\
  &   & X-Attn   & 14.53 & \\
  &   & \textbf{Concat} & \textbf{12.88} & \\
\midrule

\multirow{6}{*}{WavLM}
  & \multirow{3}{*}{$\Delta$HuBERT}
      & Weighted & 23.86 & \multirow{6}{*}{1h} \\
  &   & X-Attn   & 23.11 & \\
  &   & Concat   & 22.74 & \\
\cmidrule(lr){2-4}
  & \multirow{3}{*}{$\Delta$W2V2}
      & Weighted & 23.42 & \\
  &   & X-Attn   & 25.97 & \\
  &   & \textbf{Concat} & \textbf{21.81} & \\
\bottomrule
\end{tabular}}
\label{tab:fusion_results_myst}
\end{table}


\subsubsection{Comparison with Individual Fine-Tuned Models}

Table~\ref{tab:ft_baselines} compares the best delta fusion method (Concat) with single fine-tuned models.  
All fusions with $\Delta$W2V2 yield statistically significant improvements ($p < 0.05$) over the reference model WavLM, demonstrating the robustness of delta fusion, particularly in few-shot settings. Fusion with $\Delta$HuBERT also provides statistically significant gains in most settings. We further analyze both delta embeddings in Section 4.2 to investigate the performance gain.

\begin{table}[ht]
\centering
\vspace{-6pt}
\caption{WERs (\%) for fine-tuned SSL models and delta embedding fusion (Concat). 
Bold indicates the best result per data condition. * denotes statistical significance ($p<0.05$) compared with the reference model WavLM.}
\setlength{\tabcolsep}{7pt}
\begin{tabular}{lcccc}
\toprule
\textbf{Model} & \textbf{Full} & \textbf{10h} & \textbf{5h} & \textbf{1h} \\
\midrule
WavLM                  & 10.16 & 11.95 & 13.27 & 22.47 \\
HuBERT                 & 11.04 & 12.95 & 14.67 & 25.30 \\
W2V2                   & 10.96 & 13.47 & 15.65 & 25.97 \\
WavLM + $\Delta$HuBERT &  9.71* & \textbf{11.57*} & 12.96* & 22.74 \\
WavLM + $\Delta$W2V2   & \textbf{9.64*} & 11.61* & \textbf{12.88*} & \textbf{21.81*} \\
\bottomrule
\end{tabular}
\label{tab:ft_baselines}
\end{table}

\subsubsection{Fusion of Fine-Tuned vs.\ Delta Embeddings}

We compare fusion using fine-tuned embeddings with fusion using delta embeddings, where both approaches use the same number of parameters. As shown in Table~\ref{tab:ft_fusion_concat}, WavLM\(+\Delta\)W2V2 performs on par with WavLM\(+\)W2V2 in most cases, while achieving a 4.4\% relative WER reduction in the challenging 1\,h scenario. In contrast, WavLM\(+\Delta\)HuBERT consistently outperforms WavLM\(+\)HuBERT across all training conditions, underscoring the effectiveness of delta embedding fusion in low-resource settings.  

Notably, fusing the reference model WavLM with \(\Delta\)HuBERT yields statistically significant improvements (\(p<0.05\)) over fusion with fine-tuned HuBERT under every data condition. In the 1\,h setting, \(\Delta\)HuBERT achieves a 10\% relative WER reduction compared to fine-tuned HuBERT.  


\begin{table}[ht]
\centering
\vspace{-6pt}
\caption{WERs (\%) when fusing WavLM with HuBERT/W2V2 versus their
$\Delta$ counterparts (Concat).}
\setlength{\tabcolsep}{8pt}
\begin{tabular}{l|cc|cc}
\toprule
\textbf{Data} & \textbf{HuBERT} & \textbf{$\Delta$HuBERT} & \textbf{W2V2} & \textbf{$\Delta$W2V2} \\
\midrule
Full & 10.35 & 9.71* & 9.67 & 9.64 \\
10h  & 11.84 & 11.57* & 11.66 & 11.61 \\
5h   & 13.21 & 12.96* & 12.89 & 12.88 \\
1h   & 25.27 & 22.74* & 22.80 & 21.81* \\
\bottomrule
\end{tabular}
\label{tab:ft_fusion_concat}
\end{table}


\subsection{How do delta embeddings improve fusion performance?}

\subsubsection{Canonical Correlation Analysis}
To understand why delta embeddings improve fusion, we perform a layer-wise CCA on the MyST development set. We compute the similarity between fine-tuned models and (i) their pre-trained counterparts and (ii) their delta embeddings.  

\begin{figure}[ht]
\centering
\vspace{-6pt}
\includegraphics[width=0.38\textwidth]{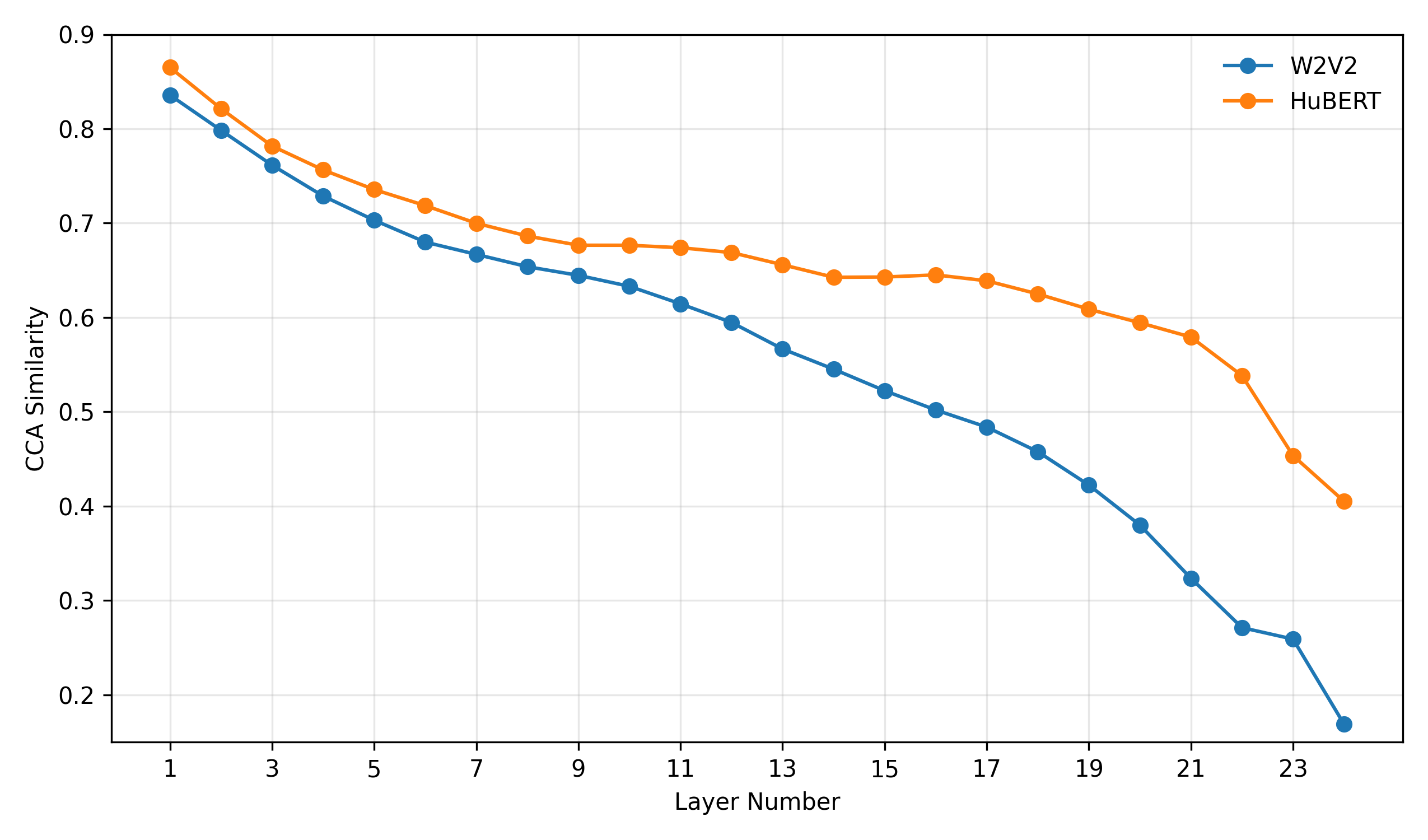}
\caption{CCA similarity between pre-trained and fine-tuned models.}
\label{fig:fig1}
\end{figure}

Figure~\ref{fig:fig1} presents the CCA similarity between pre-trained and fine-tuned representations. For both HuBERT and W2V2, similarity decreases with depth, confirming that fine-tuning primarily affects the upper transformer layers. HuBERT retains a higher similarity in deeper layers compared to W2V2, suggesting a more gradual adaptation to the task~\cite{huo2025iterative}. This observation helps explain why fusing fine-tuned WavLM and HuBERT embeddings contains greater redundancy, as the two models share more similar pretraining methodologies. In contrast, W2V2 undergoes sharper task-specific shifts during fine-tuning, making its fine-tuned embeddings nearly as effective as $\Delta$W2V2 embeddings in fusion except under extremely low-resource conditions, as shown in Table \ref{tab:ft_baselines}.

\begin{figure}[ht]
\centering
\vspace{-6pt}
\includegraphics[width=0.38\textwidth]{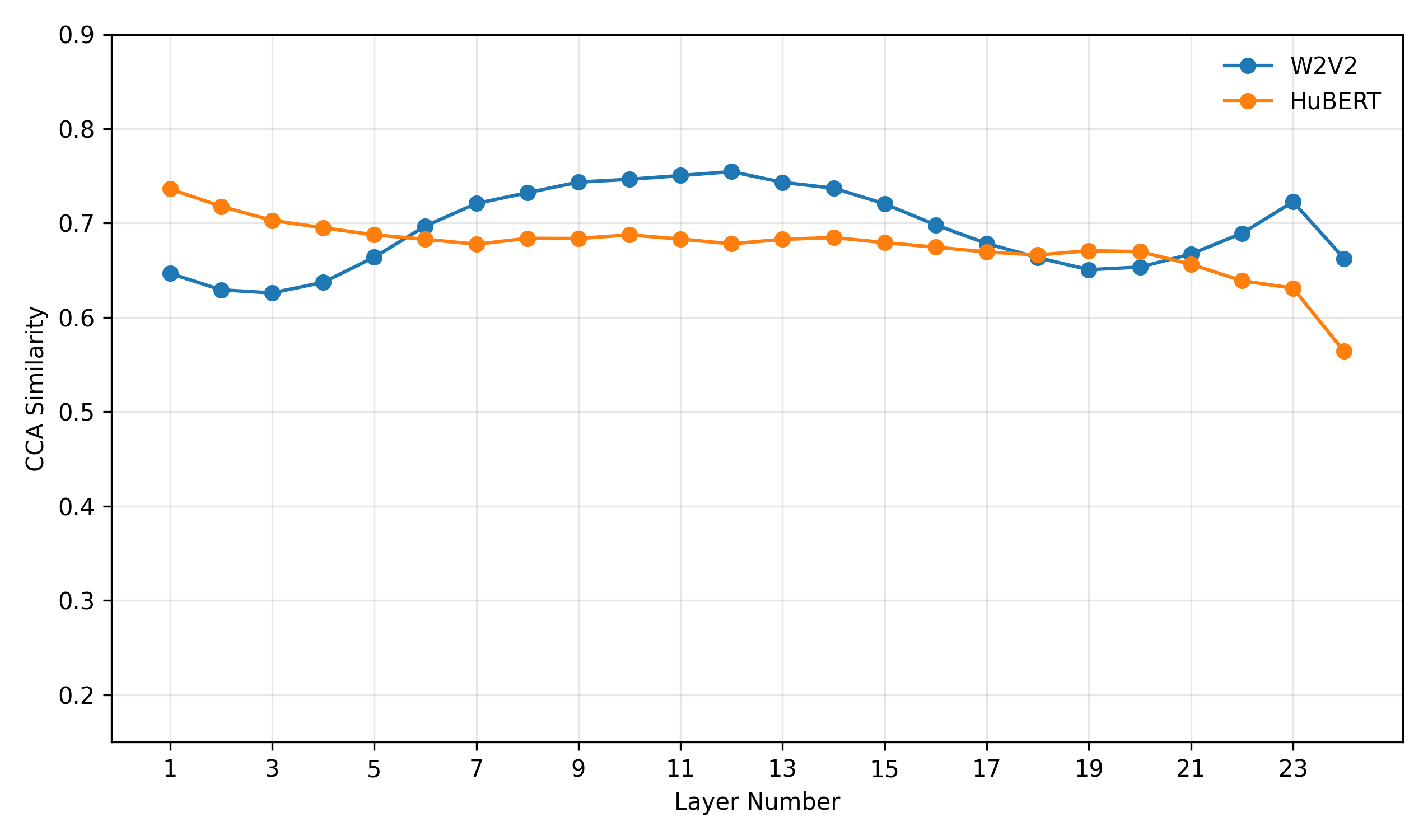}
\caption{CCA similarity between fine-tuned and $\Delta$ embeddings.}
\label{fig:fig2}
\end{figure}

Figure~\ref{fig:fig2} reports the CCA similarity between fine-tuned models and their delta embeddings. For both HuBERT and W2V2, similarity remains relatively stable across the middle layers before dropping at the final layer, suggesting that delta embeddings primarily capture task-specific shifts concentrated in the upper layers. In particular, $\Delta$HuBERT exhibits a smoother and more gradual decline, while $\Delta$W2V2 shows a sharper drop at the output layer. This contrast suggests that $\Delta$W2V2 captures stronger task-specific deviations, consistent with its observed complementarity in fusion experiments.

\subsubsection{Cross-Domain Analysis}

We further evaluate \emph{cross-domain} delta embeddings by fusing fine-tuned WavLM (MyST) with $\Delta$HuBERT and $\Delta$W2V2 from models fine-tuned on 100h LibriSpeech \cite{panayotov2015librispeech}, and compare them against \emph{in-domain} deltas fine-tuned on MyST.
As shown in Table~\ref{tab:delta_domain}, in-domain deltas consistently achieve lower WERs, indicating that delta embeddings capture task-specific information. And cross-domain deltas still outperform the WavLM baseline, aligning with prior evidence that adult speech knowledge benefits child ASR \cite{shankar2025selective}.

\begin{table}[!ht]
\centering
\vspace{-6pt}
\caption{WERs (\%) for fusion of fine-tuned WavLM (MyST) with $\Delta$ embeddings from models fine-tuned on \emph{MyST} vs.\ \emph{LibriSpeech}.}
\setlength{\tabcolsep}{8pt}
\begin{tabular}{lcc}
\toprule
\textbf{Fusion (WavLM + $\Delta$)} & \textbf{MyST} & \textbf{Libri} \\
\midrule
HuBERT & 9.71 & 10.06 \\
W2V2   & \textbf{9.64} & 9.99 \\
\bottomrule
\end{tabular}
\label{tab:delta_domain}
\end{table}

\subsubsection{Mixture of Experts Analysis}



We analyze the normalized MoE gating weights ($w_{\text{ft}} + w_{\Delta} = 1$) defined in Section~\ref{sec:moe}. Table~\ref{tab:moe_results} reports the mean weight $\overline{w}_{\text{ft}}$ per utterance. We observe that the model assigns significant weight to both embeddings, confirming their joint utility. Notably, W2V2 fusion results in a lower $\overline{w}_{\text{ft}}$ than HuBERT, which correlates with improved WER. This suggests that W2V2 offers greater complementarity to WavLM due to distinct pre-training strategies, whereas HuBERT's similar objective leads to higher redundancy.

\begin{table}[!ht]
\centering
\vspace{-6pt}
\caption{WERs (\%) for MoE results.
$\overline{w}_{\text{ft}}(x)$ is the mean MoE weight assigned to the WavLM embedding.}
\label{tab:moe_results}
\begin{tabular}{lcc}
\toprule
\textbf{Fusion Model} & $\overline{w}_{\text{ft}}(S)$ & \textbf{WER} \\
\midrule
WavLM + $\Delta$ HuBERT & 0.72 & 9.83 \\
WavLM + $\Delta$ W2V2   & 0.65 & 9.70 \\
\bottomrule
\end{tabular}
\end{table}




\section{Conclusion}
\label{sec:conclusion}

This paper proposes the use of delta embeddings for fusing SSL representations in child ASR.  
On the MyST corpus, concatenating fine-tuned WavLM with
$\Delta$W2V2 consistently reduced WER compared to single-model baselines and alternative fusion strategies. Notably, fusing WavLM with $\Delta$W2V2 achieved a WER of 9.64, establishing a new state-of-the-art result among SSL models on the MyST corpus. 
Compared to fusion with fine-tuned embeddings, delta embeddings yielded particularly strong gains in the extremely low-resource 1\,h setting, achieving a 10\% relative WER reduction for HuBERT and a 4.4\% reduction for W2V2.

CCA and cross-domain analyses demonstrated that delta embeddings isolate
task-specific shifts,
explaining their effectiveness in fusion.  
Overall, delta embeddings provide a simple yet effective mechanism for
multi-model fusion in child ASR. Future work will evaluate delta embedding fusion in other low-resource domains.


\section{Acknowledgments}
This research is supported in part by the IES, U.S. Department of Education (DoE) through Grant R305C240046 to the U. at Buffalo, and the NSF. The opinions expressed are those of the authors and do not represent views of the IES, DoE, or the NSF.

\bibliographystyle{IEEEbib}
\bibliography{strings,refs}

@article{mohamed2022self,
  title={Self-supervised speech representation learning: A review},
  author={A. Mohamed and others},
  journal={IEEE Journal of Selected Topics in Signal Processing},
 volume={16},
  number={6},
  pages={1179--1210},
  year={2022},
}

@inproceedings{fan2024benchmarking,
  author={R. Fan and others},
  title={Benchmarking Children's ASR with Supervised and Self-supervised Speech Foundation Models},
  booktitle={Proc. Interspeech},
  year={2024}
}

@inproceedings{gerosa2006analyzing,
  title={Analyzing children's speech: An acoustic study of consonants and consonant-vowel transition},
  author={M. Gerosa and others},
  booktitle={2006 IEEE International Conference on Acoustics Speech and Signal Processing Proceedings},
  volume={1},
  year={2006},

}

@article{lee1999acoustics,
  title={Acoustics of children’s speech: Developmental changes of temporal and spectral parameters},
  author={S. Lee and others},
  journal={The Journal of the Acoustical Society of America},
  volume={105},
  number={3},
  pages={1455--1468},
  year={1999},
  publisher={Acoustical Society of America}
}

@inproceedings{yeung2018difficulties,
  title = {On the difficulties of automatic speech recognition for kindergarten-aged children},
  author = {G. Yeung and A. Alwan},
  booktitle = {Proc. Interspeech},
  year = {2018},
}

@inproceedings{dutta2022challenges,
  title = {Challenges remain in building {ASR} for spontaneous preschool children speech in naturalistic educational environments},
  author = {Dutta, S. and others},
  booktitle = {Proc. Interspeech},
  year = {2022},
}

@article{ying2025benchmarking,
  title={Benchmarking Training Paradigms, Dataset Composition, and Model Scaling for Child ASR in ESPnet},
  author={A. Ying and others},
  journal={arXiv preprint arXiv:2508.16576},
  year={2025}
}

@inproceedings{attia2025cpt,
  title={Cpt-boosted wav2vec2. 0: Towards noise robust speech recognition for classroom environments},
  author={A. A. Attia and others},
  booktitle={Proc. ICASSP},
  year={2025},

}

@inproceedings{carvalho2025ac,
  title={AC-Mix: Self-Supervised Adaptation for Low-Resource Automatic Speech Recognition using Agnostic Contrastive Mixup},
  author={C. Carvalho and A. Abad},
  booktitle={Proc. ICASSP},
  year={2025},

}

@inproceedings{sinha2025beyond,
  title={Beyond Traditional Speech Modifications: Utilizing Self Supervised Features for Enhanced Zero-Shot Children ASR},
  author={A. Sinha and others},
  booktitle={Proc. Interspeech},
  year={2025}
}

@inproceedings{horii2025children,
  title={Why is children’s ASR so difficult? Analyzing children’s phonological error patterns using SSL-based phoneme recognizers},
  author={K. Horii and others},
  booktitle={Proc. Interspeech},
  year={2025}
}

@inproceedings{medin2024self,
  title={Self-Supervised Models for Phoneme Recognition: Applications in Children's Speech for Reading Learning},
  author={L. Medin and others},
  booktitle={Proc. Interspeech},
  year={2024},
}

@inproceedings{li2023towards,
  title={Towards Robust Family-Infant Audio Analysis Based on Unsupervised Pretraining of Wav2vec 2.0 on Large-Scale Unlabeled Family Audio},
  author={J. Li and others},
  booktitle={Proc. Interspeech},
  year={2023}
}

@article{sinha2025can,
  title={Can Layer-wise SSL Features Improve Zero-Shot ASR Performance for Children's Speech?},
  author={A. Sinha and others},
  journal={IEEE Signal Processing Letters},
  year={2025},
}

@inproceedings{radford2023robust,
  title={Robust speech recognition via large-scale weak supervision},
  author={A. Radford and others},
  booktitle={Proc. ICML},
  year={2023},
}

@article{baevski2020wav2vec,
  title={Wav2vec 2.0: A framework for self-supervised learning of speech representations},
  author={A. Baevski and others},
  journal={Advances in Neural Information Processing Systems},
  volume={33},
  pages={12449--12460},
  year={2020}
}

@article{hsu2021hubert,
  title={Hubert: Self-supervised speech representation learning by masked prediction of hidden units},
  author={W.-N. Hsu and others},
  journal={IEEE/ACM Transactions on Audio, Speech, and Language Processing},
  volume={29},
  pages={3451--3460},
  year={2021},
}

@article{chen2022wavlm,
  title={Wavlm: Large-scale self-supervised pre-training for full stack speech processing},
  author={S. Chen and others},
  journal={IEEE Journal of Selected Topics in Signal Processing},
  volume={16},
  number={6},
  pages={1505--1518},
  year={2022},
}

@inproceedings{baevski2022data2vec,
  title     = {data2vec: A general framework for self-supervised learning in speech, vision and language},
  author    = {A. Baevski and others},
  booktitle = {Proc. ICML},
  year      = {2022},
}

@inproceedings{tang2022exploring,
  title={Exploring Effective Fusion Algorithms for Speech Based Self-Supervised Learning Models},
  author={C. Tang and others},
  booktitle={Proc. NCMMSC},
  year={2022}
}

@inproceedings{chen2022fearless,
  title={FeaRLESS: Feature Refinement Loss for Ensembling Self-Supervised Learning Features in Robust End-to-end Speech Recognition},
  author={S.-J. Chen and others},
  booktitle={Proc. Interspeech},
  year={2022}
}

@inproceedings{srivastava2023effuse,
  author={T. Srivastava and others},
  title={{{EFFUSE:} Efficient Self-Supervised Feature Fusion for {E2E} {ASR}
                  in Low Resource and Multilingual Scenarios}},
  year={2024},
  booktitle={Proc. Interspeech},
}

@inproceedings{chiu2024learnable,
  title     = {Learnable Layer Selection and Model Fusion for Speech Self-Supervised Learning Models},
  author    = {S.-C. Chiu and others},
  booktitle = {Proc. Interspeech},
  year      = {2024}
}

@inproceedings{arunkumar2022investigation,
  title     = {Investigation of ensemble features of self-supervised pretrained models for automatic speech recognition},
  author    = {A. Arunkumar and V. N. Sukhadia and S. Umesh},
  booktitle = {Proc. Interspeech},
  year      = {2022}
}

@inproceedings{ramesh2024task,
  title     = {Task vector algebra for ASR models},
  author    = {G. Ramesh and K. Audhkhasi and B. Ramabhadran},
  booktitle = {Proc. ICASSP},
  year      = {2024},
}

@inproceedings{shankar2025selective,
  title     = {Selective Attention Merging for low resource tasks: A case study of child ASR},
  author    = {N. Shankar and Z. Wang and E. Eren and A. Alwan},
  booktitle = {Proc. ICASSP},
  year      = {2025},
}

@inproceedings{ducorroy2025robust,
  title     = {Robust fine-tuning of speech recognition models via model merging: application to disordered speech},
  author    = {Ducorroy, Alexandre and Riad, Rachid},
  booktitle = {Proc. Interspeech},
  year      = {2025},
}

@inproceedings{zhang2019delta,
  title={Delta Embedding Learning},
  author={X. Zhang and J. Wu and D. Dou},
  booktitle={Proc. ACL},
  year={2019}
}

@inproceedings{pasad2021layer,
  title     = {Layer-wise analysis of a self-supervised speech representation model},
  author    = {A. Pasad and J.-C. Chou and K. Livescu},
  booktitle = {Proc. ASRU},
  year      = {2021},
}

@article{jacobs1991moe,
  author  = {R. Jacobs and M. Jordan and S. Nowlan and G. Hinton},
  title   = {Adaptive mixtures of local experts},
  journal = {Neural Computation},
  year    = {1991},
  volume  = {3},
  number  = {1},
  pages   = {79--87},
}

@inproceedings{berrebbi2022combining,
  title={Combining Spectral and Self-Supervised Features for Low Resource Speech Recognition and Translation},
  author={D. Berrebbi and others},
  booktitle={Proc. Interspeech},
  year={2022}
}

@article{pasad2024self,
  title   = {What do self-supervised speech models know about words?},
  author  = {A. Pasad and others},
  journal = {Transactions of the Association for Computational Linguistics},
  volume  = {12},
  pages   = {372--391},
  year    = {2024}
}

@inproceedings{pasad2023comparative,
  title     = {Comparative layer-wise analysis of self-supervised speech models},
  author    = {A. Pasad and others},
  booktitle = {Proc. ICASSP},
  year      = {2023},
}

@incollection{hotelling1992relations,
  title     = {Relations between two sets of variates},
  author    = {H. Hotelling},
  booktitle = {Breakthroughs in Statistics: Methodology and Distribution},
  pages     = {162--190},
  year      = {1992},
  publisher = {Springer}
}

@article{morcos2018insights,
  title   = {Insights on representational similarity in neural networks with canonical correlation},
  author  = {A. Morcos and others},
  journal = {Advances in Neural Information Processing Systems},
  volume  = {31},
  year    = {2018}
}

@article{ward2011my,
  title   = {My science tutor: A conversational multimedia virtual tutor for elementary school science},
  author  = {W. Ward and others},
  journal = {ACM Transactions on Speech and Language Processing (TSLP)},
  volume  = {7},
  number  = {4},
  pages   = {1--29},
  year    = {2011}
}

@inproceedings{attia2023kid,
  title={Kid-whisper: Towards bridging the performance gap in automatic speech recognition for children vs. adults},
  author={A. Attia and others},
  booktitle={Proc. AAAI/ACM Conference on AI, Ethics, and Society},
  year={2024}
}

@inproceedings{wolf2020transformers,
  title     = {Transformers: State-of-the-art natural language processing},
  author    = {T. Wolf and others},
  booktitle = {Proc. EMNLP: System Demonstrations},
  year      = {2020}
}

@inproceedings{huo2025iterative,
  title={Iterative Refinement, Not Training Objective, Makes HuBERT Behave Differently from wav2vec 2.0},
  author={R. Huo and E. Dunbar},
  booktitle={Proc. Interspeech},
  year={2025}
}

@inproceedings{panayotov2015librispeech,
  title     = {LibriSpeech: An ASR corpus based on public domain audio books},
  author    = {V. Panayotov and others},
  booktitle = {Proc. ICASSP},
  year      = {2015},
}

\end{document}